Hellenic Complex Systems Laboratory

# Design of Statistical Quality Control Procedures Using Genetic Algorithms

Technical Report II

Aristides T. Hatjimihail and Theophanes T. Hatjimihail

Hellenic Complex Systems Laboratory,

P.O. Box 56,

GR-66100 Drama, Greece

www.hcsl.com

# Design of Statistical Quality Control Procedures Using Genetic Algorithms

**Aristides T. Hatjimihail, Theophanes T. Hatjimihail**

Hellenic Complex Systems Laboratory

P.O. Box 56, GR-661 00 Drama, Greece.

## Abstract

In general, we can not use algebraic or enumerative methods to optimize a quality control (QC) procedure so as to detect the critical random and systematic analytical errors with stated probabilities, while the probability for false rejection is minimum. Genetic algorithms (GAs) offer an alternative, as they do not require knowledge of the objective function to be optimized and search through large parameter spaces quickly. To explore the application of GAs in statistical QC, we have developed an interactive GAs based computer program that designs a novel near optimal QC procedure, given an analytical process. The program uses the deterministic crowding algorithm. An illustrative application of the program suggests that it has the potential to design QC procedures that are significantly better than 45 alternative ones that are used in the clinical laboratories

## 1 INTRODUCTION

According to the International Federation of Clinical Chemistry (IFCC), "the most important purpose of clinical chemistry is to measure, in body fluids and tissues of individual patients, those substances which are relevant for the understanding, prevention, diagnosis, or treatment of disease" (The International Federation of Clinical Chemistry Committee on Standards, 1975).

The improvement of the performance of the analytical procedures has been the primary goal of the statistical (QC), in clinical chemistry. The performance of a measurement procedure is generally described in terms of precision and accuracy (Westgard, Barry, 1986). Precision is the agreement of replicate measurements, while imprecision or *random error* is the standard deviation of the results of a set of replicate measurements. Accuracy is the agreement between the best estimate of a quantity and its true value, while inaccuracy or *systematic error* is the difference between the mean of a set of replicate measurements and the true value (The International Federation of Clinical Chemistry Committee on Standards, 1975). As *critical errors* are considered the maximal allowable random and systematic errors "defined in such a way that an upper bound has been set on the (clinical) type I error" (Linnet, 1989). The clinical type I error is the probability for rejection of the true hypothesis that there is no significant change of an analyte of a patient. The hypothesis is rejected when the observed change of the analyte is greater than the *maximum medically allowable total error.*

Statistical QC in clinical chemistry is achieved mainly through the repetitive analysis of stable control materials over long periods of time. One or more specimens of control materials, at one or more levels of analyte concentration, are analysed in parallel with the patients' specimens, during each analytical run. Statistical *QC procedures* are applied upon the data to detect the introduction of random or systematic error greater than the respective critical errors. The analytical process is then considered out of control, and the analytical run is rejected. A *QC procedure* is a Boolean proposition, composed from one or more statistical decision rules. Each decision rule is applied upon a sample of control measurements and includes the calculation of a *statistic* and the comparison of that with a *control limit*.

Alternative QC procedures can be designed to test statistically the null hypothesis (the analytical process is in control) against the alternative (the analytical process is out of control). When a true null hypothesis is rejected, a type I error is committed. We have then a false rejection of an analytical run. The probability of a type I error is called

*probability for false rejection*. When a false null hypothesis is accepted, a type II error is committed. We fail then to detect a significant change in the distribution of error in the analytical process. The probability for rejection of a false null hypothesis is called *probability for error detection*. An *optimal QC procedure* should have stated probabilities for detecting the critical random and systematic errors, while the probability for false rejection is minimum.

The probabilities for random and systematic error detection and for false rejection are estimated by computer simulation (Hatjimihail, 1992; Westgard, Groth, 1981), because the algebraic definition of the functions relating the probability for error detection with the size of the analytical error is very complex, if possible, in most cases. Therefore we can not use algebraic methods to optimize the QC procedures. Usage of enumerative methods would be very tedious, especially with multi-rule procedures, as the number of the points of the parameter space to be searched grows exponentially with the number of the parameters to be optimized. Optimization methods based on the GAs offer an appealing alternative as they are robust search algorithms, that do not require knowledge of the objective function and search through large spaces quickly. GAs have been derived from the processes of the molecular biology of the gene and the evolution of life. Their operators, cross-over, mutation, and reproduction, are isomorphic with the synonymous biological processes. GAs have been successfully used to solve a variety of complex optimization problems (Alander, 1994; Davis, 1991), including the optimization of QC procedures (Hatjimihail, 1993). Furthermore, the complexity of the design process of novel QC procedures is obviously greater than the complexity of the optimization of predefined ones. The classifier systems (Holland, 1992; Goldberg, 1989) and the genetic programming paradigm (Koza, 1994) have shown us that GAs can be used for tasks as complex as the program induction. Indeed, GAs have been used for the design of statistical QC procedures, as well (Hatjimihail, 1993), although the previously described program has not designed statistical QC procedures significantly better than the commonly used ones.

To explore the application of the GAs in statistical QC, we have developed a GAs based, interactive, microcomputer program for the design of optimized QC procedures. The user defines the maximum number of the rules of the QC procedure, the parameters of the analytical process, and the parameters of the genetic search. Then the program defines a near optimal QC procedure, using any of four generic QC rules (see Appendix).

# 2 MATERIALS AND METHODS

## 2.1 THE PROGRAM

The program is written in Pascal, for computers with an Intel 80x86 central processing unit (Intel Corporation, U.S.A.), under MS-DOS or Windows (Microsoft Corporation, U.S.A.). It includes the following units:

### 2.1.1 The Rules Unit

The rules used by the program belong to four classes: Single Value Rules, Range Rules, Mean Rules, and Standard Deviation Rules (see Appendix). The user defines the maximum number *a* of the rules of the QC procedure. The procedure to be designed by the program can be denoted as:

$Q_1(n_1,max_1)$ # $Q_2(n_2,max_2)$ #... # $Q_q(n_q,max_q)$,
where $1 \leq a \leq q$.

The following notation is used: Let *n* be the sample size, that is the number of the measurements the QC rule is applied upon, and $x(SD)$ the decision limit. Then $S(n,x)$ denotes a Single Value Rule, $R(n,x)$ a Range Rule, $M(n,x)$ a Mean Rule, $D(n, x)$ a Standard Deviation Rule and $Q(n,x)$ any rule from the above. A rule is true if the respective statistic is greater than *x*. The symbol # denotes either the operator AND or the operator OR.

The priority of the operators is optimized by the program, therefore the designed procedure may have any parentheses. The rules are applied across runs and across the levels of the control measurements.

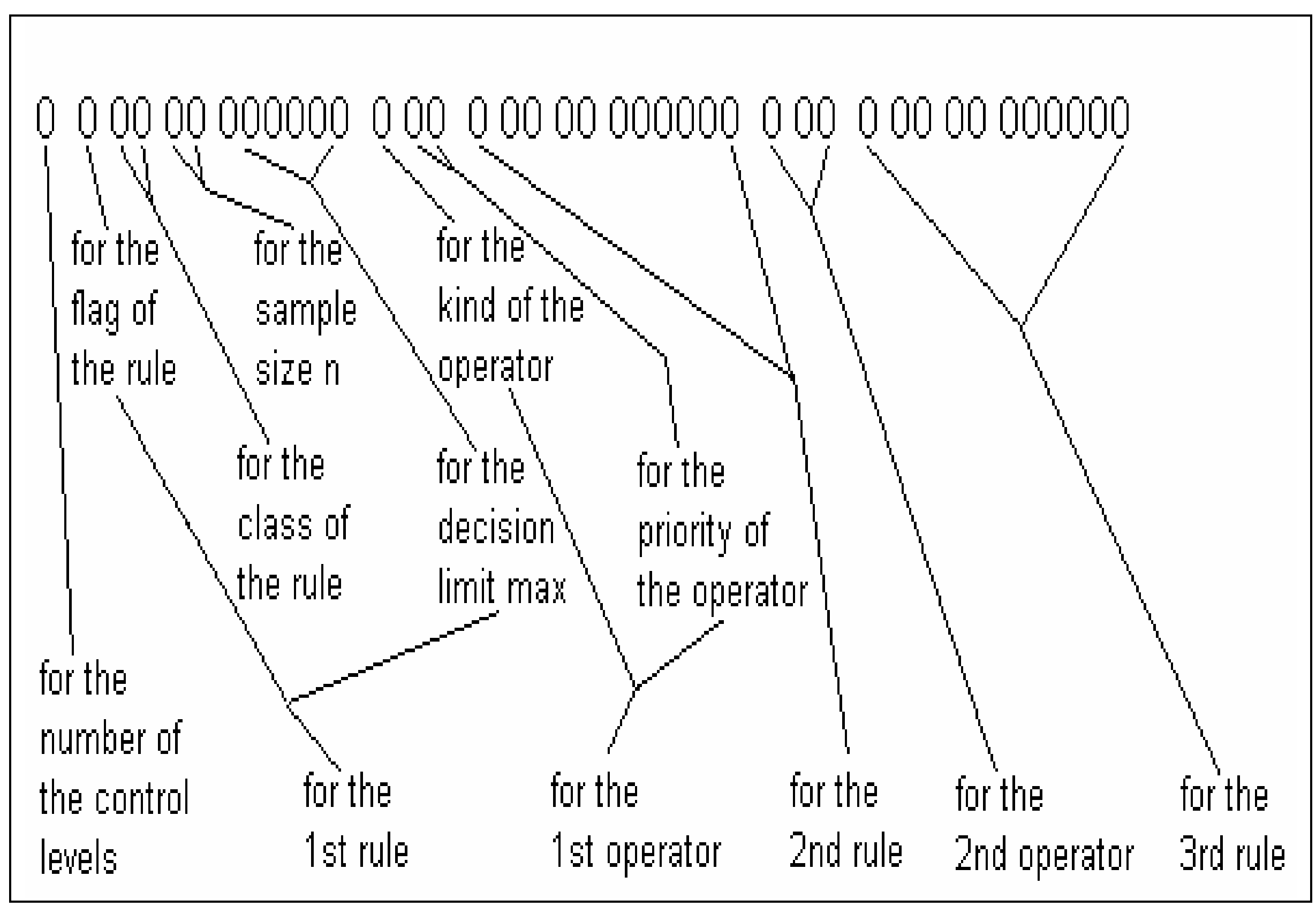

Figure 1: The Bit Mapping of Each String of the Application of the Program

### 2.1.2 The Genetic Algorithms' Unit

Each QC rule is coded as a substring of eleven bits. One bit codes a flag; if the flag equals one, then the rule and the operator following the rule are included in the procedure, else they are not. Two bits code the class of the rule, two bits the sample size $n$, and six bits the decision limit *max*. Each operator is coded as a substring of three bits. One bit codes the kind of the operator (AND or OR), and two bits the priority of the operation. A QC procedure is composed of $a$ rules and $(a-1)$ operators, where $1 \leq a \leq q$, and it is coded by a string of $10q + 3(q - 1)$ bits.

The number of the control measurements per level (up to four) and the number of their levels (up to two) are either user defined or optimized. In addition, the string may includeone bit coding the number of the levels of the control measurements (up to two) and up to two bits coding the number of the control measurements per level (up to four), when these parameters are optimizedThe bounds of the parameters to be optimized are predefined (see Appendix). The parameters of the genetic search are user defined. The fitness $f$ of the translated strings is calculated by the following objective function:

$$f = \sqrt{w_{re} (P_{re} - P_{re_c})^2 + w_{se} (P_{se} - P_{se_c})^2 + w_{fr} P_{fr}^2} \quad [1]$$

where $P_{re}$ and $P_{se}$ are the probabilities for random andsystematic critical error detection, and $P_{fr}$ the probabilityfor false rejection of the QC procedure defined by the translated string, $P_{re_c}$and $P_{se_c}$ the stated probabilities for critical random and systematic error detection, and $w_{re}$, $w_{se}$, ,$w_{fr}$, weighing factors. The probabilities for critical random and systematic error detection and for false rejection are estimated by simulation. The weighing factors $w_{re}$, $w_{se}$, ,$w_{fr}$, of the objective function [1] are set to 1. The objective function f is minimized.

The strings are evaluated, and then are reproduced according to the deterministic crowding algorithm (Mahfoud, 1992), that preserves multimodal diversity. When the fitness of the two competing strings is equal, the string that codes the procedure with the smaller number of operators is selected.

### 2.1.3 The Simulator Unit

The program simulates 1000 control measurements at each level of the control measurements, in consecutive runs, and applies to them successively the designed QC procedure. To estimate the probability for critical random and systematic error detection, the program introduces into the simulated control measurements the critical random and systematic errors respectively. The critical errors are calculated as it has been described (Hatjimihail, 1993). The program assumes a normal distribution of error. The unit is similar to that described previously (Hatjimihail, 1992), modified according to the guidelines of Parvin (Parvin, 1991).

### 2.1.4 The Random Numbers Generator Unit

The simulated control measurements are random normal deviates. The random number generator is based on the FORTRAN code of Marse and Roberts (Law, 1991; Marse, Roberts,1983).

## 3 RESULTS

### 3.1 APPLICATION OF THE PROGRAM

To illustrate the process of the design of an optimized QC procedure we have applied the program to data obtained from a clinical laboratory. The SD and the bias of an analytical method for the measurement of sodium in serum were 0.67 meq/lit and 0.1 meq/lit respectively (Koch et al, 1990). The maximum allowable analytical error for the measurement of sodium in serum is 4.0 meq/lit (Koch et al, 1990). The upper bound of the clinical type I error was set to 0.01. Ignoring the pre analytical and biological variation, and assuming that we analyse only one sample per test, the critical random and systematic errors were calculated equal to 2.313 SD and 3.495 SD respectively. The stated probabilities for critical random and systematic error detection were set to 0.5 and 1.0 respectively.

The population of the strings was set to 600. The probability for crossover was set to 1. The probability for mutation was set to 0 during the first 50 generations, then it was set to 0.0005. The genetic search continued for 100 generations.

A QC procedure with up to three rules was designed, to be applied to up to two levels of control measurements, procedure was coded by a string of $11 \times 3+3(3-1)+1=40$ bits. Each string was ordered as it is shown in the Figure 1. Although the parameter space includes more than $10^{12}$ points, there are 2,612,585,344 different possible solutions, because of the existing symmetries. Ignoring the number of the levels of the control measurements, and the parameters $n_i$ and $max_i$ of the rules, there are 184 different Boolean propositions based on the four generic rules. Obviously the problem is multimodal.

The designed QC procedures were compared with a library of 45 alternative QC procedures[1], including the QC procedures that are commonly used in the clinical laboratories. The alternative QC procedures are composed from single value, range, mean and/or standard deviation rules. Among these QC procedures are included the following, according to Westgard's notation (Westgard, Barry, 1986):

1. The Westgard procedure (Westgard et al, 1981).

2. $1_{\nu S}$ , where $\nu=2.0+0.1k$, $k=0,1,2..,20.999$

3. $1_{3.0s} / 2_{2.0s} / R_{4.0s}$

To compare the designed QC procedures with the alternative ones, we used a version of the previously described simulation program (Hatjimihail, 1992), with the random number generator of Marse and Roberts and the simulator unit of this program. Twenty one simulation runs were performed for each procedure, assuming one and then two levels of control measurements, using twenty one different sets of series of simulated control measurements. The probabilities for critical error detection, and for false rejection were estimated for all the QC procedures. For the comparison of the QC procedures we used a function similar to [1], that is the function:

$$f_1 = \sqrt{\Delta P_{re}^{\ 2} + \Delta P_{se}^{\ 2} + P_{fr}^{\ 2}} \qquad [2]$$

where $\Delta P_{re} = P_{re} - 0.5$ if $P_{re} < 0.5$ , else $\Delta P_{re} = 0$,

and $\Delta P_{se} = P_{se} - 1$.

The $f_1$ is considered optimum when it is minimum.

These are the fittest procedures, designed by the program, during each of five randomly selected consecutive runs, and the respective probabilities for the critical random

---

[1] The library of the 45 alternative QC procedures is available from the authors.

and systematic error detection, and for false rejection and the objective function f values:

1. R(2,4.3) OR M(2,1.9), designed during the 13rd generation, with:

$P_{rec} = 0.489$, $P_{sec} = 0.991$, $P_{fr} = 0.019$, and f=0.02373.

2. S(1,3.2) OR R(4,4.6) OR M(2,1.9), designed during the 81st generation, with:

$P_{rec} = 0.489$, $P_{sec} = 0.991$, $P_{fr} = 0.017$, and f=0.02216.

3. ( S(1,2.2) AND M(2,1.9)) OR R(4,4.3), designed during the 39th generation, with:

$P_{rec} = 0.495$, $P_{sec} = 0.988$, $P_{fr} = 0.019$, and f=0.02302.

4. S(1,2.7) OR M(2,1.9), designed during the 57th generation, with:

$P_{rec}= 0.492$, $P_{sec}= 0.992$, $P_{fr} = 0.022$, and f = 0.02474

5. S(1,1.9) AND (R(4,4.2) OR M(2,1.9)), designed during the 42nd generation, with:

$P_{rec} = 0.504$, $P_{sec} = 0.990$, $P_{fr} = 0.020$, and f=0.02272.

All the QC procedures were designed to be applied upon two levels of control measurements. The Table 1 presents the mean and the SD of the estimated probabilities for the critical errors detection and for false rejection, and of the values of the function $f_1$, of the designed QC procedures and of the five best procedures of the library. The procedures were sorted by the value of the function $f_1$. The differences between the values of the function $f_1$ of each of the three best procedures designed by the program, and of each of the 45 procedures of the library, are statistically significant [sign test, $p<0.05$ (Daniel, 1987)].

## 4 DISCUSSION

Traditionally, statistical QC has been a compromise between the requirements for high accuracy and precision and for low false rejections (Duncan, 1986). The concept of the medical allowable analytical error (Linnet, 1989; Westgard, Barry, 1986) offered objective criteria for the evaluation of the alternative QC procedures, in clinical chemistry. Nevertheless, the design of the QC procedures and the definition of their parameters has been based mainly on the insight of the researchers (Hatjimihail, 1992; Westgard et al, 1981; Westgard, Barry, 1986), who were presumably using a trial and error process to define them. On the other hand, most clinical laboratories use one or more predefined QC procedures, obtained from libraries of commonly used ones.

GAs based programs for the design of optimized statistical QC procedures illustrate the computational innovation that GAs are generating by combining partial solutions. This kind of methodology is strictly inductive when compared to other search methods, which are deductive (Krishnakumar, Goldberg, 1992). Considering our problem, GAs have generated novel QC procedures, significantly better than the commonly used ones (See Table 1). Up to our knowledge, this paper describes the first GAs based design of QC procedures, that are significantly better than those commonly used by the clinical laboratories. Furthermore, this methodology can be used for the design of QC procedures not only in clinical chemistry, but in any area of the statistical QC.

We have applied the deterministic crowding algorithm as a selection scheme because it preserves multimodal diversity (Mahfoud, 1992).

Further research can be done in the following areas:

1. The study of the topology of the fitness landscape.
2. The application of other GAs (Goldberg, Deb, Horn, 1992; Goldberg et al, 1993) to the design of statistical QC.
3. The implementation of alternative codings for the mapping of the QC procedures, such as the Gray coding and the Wolfram's representation of the Boolean propositions (Wolfram, 1994).
4. The calculation of the initial population of the strings (Goldberg, Deb, Clark, 1992).
5. The application of alternative methods to the analysis of the statistical QC data (Packard, 1990).
6. The application of alternative optimization methods to statistical QC.

We are working in these directions. We have already applied hill climbing methods to design optimized QC procedures, using the coding we have described. The preliminary results suggest that the GAs outperform these methods, because the fitness landscape is massively multimodal.

The design of novel optimized QC procedures deserves attention because of the obvious benefits from improving statistical QC. These procedures could detect the critical errors with stated probabilities, while the probability for false rejection would be minimum.

## 4.1 CONCLUSION

The application we have described suggests that GAs have the potential to be used as a powerful, robust research tool for the inductive design of novel, optimized QC procedures. These procedures can be significantly better than the QC procedures that have been designed so far.

**Table 1**. The Mean Probabilities for Critical Random ($P_{re}$) and Systematic ($P_{se}$) Error Detection, and for False Rejection ($P_{fr}$), and the Mean Values of the Function $f_1$

| **QUALITY CONTROL PROCEDURE** | | $\mathbf{P_{re}}$ | $\mathbf{P_{se}}$ | $\mathbf{P_{fr}}$ | $\mathbf{f_1}$ |
|---|---|---|---|---|---|
| S(1,1.9) AND (R(4,4.2) OR M(2,1.9)) [a,b] | **Mean** | 0.4961 | 0.9898 | 0.0240 | 0.0272 |
| | **S.D.** | 0.0069 | 0.0023 | 0.0033 | 0.0035 |
| S(1,2.7) OR M(2,1.9) [a,b] | **Mean** | 0.4895 | 0.9906 | 0.0250 | 0.0295 |
| | **S.D.** | 0.0074 | 0.0021 | 0.0038 | 0.0049 |
| ( S(1,2.2) AND M(2,1.9)) OR R(4,4.3) [a,b] | **Mean** | 0.4846 | 0.9871 | 0.0218 | 0.0300 |
| | **S.D.** | 0.0054 | 0.0032 | 0.0033 | 0.0050 |
| S(1,3.2) OR R(4,4.6) OR M(2,1.9) [a] | **Mean** | 0.4798 | 0.9906 | 0.0212 | 0.0313 |
| | **S.D.** | 0.0075 | 0.0027 | 0.0026 | 0.0061 |
| $1_{2.5s}/2_{2.0s}/R_{4s}$ [c] | **Mean** | 0.5028 | 0.9822 | 0.0270 | 0.0327 |
| | **S.D.** | 0.0078 | 0.0036 | 0.0035 | 0.0040 |
| $1_{2.5s}/2_{2.0s}$ [c] | **Mean** | 0.4925 | 0.9822 | 0.0257 | 0.0330 |
| | **S.D.** | 0.0079 | 0.0036 | 0.0035 | 0.0045 |
| $1_{2.5s}/2_{2.0s}/4_{1s}$ [c] | **Mean** | 0.4971 | 0.9825 | 0.0274 | 0.0334 |
| | **S.D.** | 0.0081 | 0.0036 | 0.0039 | 0.0045 |
| $1_{2.5s}/2_{2.0s}/R_{4s}/4_{1s}$ [c] | **Mean** | 0.5077 | 0.9825 | 0.0288 | 0.0338 |
| | **S.D.** | 0.0076 | 0.0036 | 0.0041 | 0.0044 |
| R(2,4.3) OR M(2,1.9) [a] | **Mean** | 0.4717 | 0.9901 | 0.0212 | 0.0371 |
| | **S.D.** | 0.0072 | 0.0027 | 0.0026 | 0.0063 |
| $1_{2.4s}$ [c] | **Mean** | 0.5063 | 0.9798 | 0.0312 | 0.0375 |
| | **S.D.** | 0.0077 | 0.0046 | 0.0032 | 0.0034 |

*(a)* A QC procedure designed by the program.

*(b)* A QC procedure significantly better than the procedures of the library.

*(c)* A QC procedure of the library.

All the procedures are applied upon one control measurement per level and two levels of control measurements.

## Appendix

If *SD* and *m* are the standard deviation and mean of the control measurements, when the analytical process is in control, *n* is the sample size, and *max* is the decision limit, then the generic rules of the program are:

1. Single Value Rule: The absolute values of the last *n* control measurements are greater than *m* plus (*max*)(*SD*).

2. Range Rule: The range of the last *n* control measurements is greater than (*max*)(*SD*).

3. Mean Rule: The absolute value of the difference between the mean of the last *n* control measurements and *m* is greater than (*max*)(*SD*).

4. Standard Deviation Rule: The standard deviation of the last *n* control measurements is greater than (*max*)(*SD*).

Bounds of the parameters of the rule 1: $1 \le n \le 4$, and $0 \le max \le 6.3$. Bounds of the parameters of the rules 2,3, and 4: $2 \le n \le 4$ , and $0 \le max \le 6.3$. Since *max* is coded by a 6 bit substring the accuracy of the calculation of *max* is $\frac{6.3}{2^6 - 1} = 0.1$.

A concatenated, mapped, unsigned, binary coding is used for the coding of the parameters.

## Acknowledgments

We wish to thank David E. Goldberg, for his continuing support and encouragement.

## References

Alander J. An Indexed bibliography of genetic algorithms: Years 1957-1993. Vaasa, Finland: University of Vaasa, 1994:1-196.

Daniel WW. Biostatistics: A foundation for analysis in the health sciences. New York: John Wiley 1987:583-636.

Davis L (ed). Handbook of genetic algorithms. New York: van Nostrand Reinhold 1991:1-385.

Duncan AJ. Quality control and industrial statistics. Homewood, Illinois: Irwin, 1986:417-460, 476-512.

Goldberg DE. Genetic algorithms in search, optimization, and machine learning. Reading, MA. Addison-Wesley, 1989:217-307.

Goldberg DE, Deb K, Clark JH. Genetic algorithms, noise, and the sizing of populations. Complex Systems, 1992;6:333-62.

Goldberg DE, Deb K, Horn J. Massive multimodality, deception, and genetic algorithms. In Männer R, Manderick B (eds). Parallel problem solving from nature, 2. Amsterdam: Elsevier Science Publishers B.V. 1992:27-36.

Goldberg DE et al. Rapid, accurate optimization of difficult problems using fast messy genetic algorithms. In Forrest S (ed). Fifth International Conference on Genetic Algorithms. San Mateo: Morgan-Kaufmann 1993:56-64.

Hatjimihail AT. A tool for the design and evaluation of alternative quality control procedures. Clin Chem 1992;38:204-10.

Hatjimihail AT. Genetic algorithms-based design and optimization of statistical quality control procedures. Clin Chem 1993;39:1972-8

Holland JH. Adaptation in natural and artificial systems.

Cambridge, Massachusetts: The MIT Press 1992:1-19, 66-88.

Koch DD et al. Selection of medically useful quality control procedures for individual tests done in a multitest analytical system. Clin Chem 1990;36:230-3.

Koza JR. Genetic programming II. Cambridge, MA: MIT Press, 1994:1-746.

Krishnakumar K, Goldberg DE. Control system optimi zation using genetic algorithms. J Guid Control Dyn 1992;15:735-40.

Law AM, Kelton WD. Simulation modelling and analysis. New York: McGraw-Hill, 1991:420-461.

Linnet K. Choosing quality-control systems to detect maximum clinically allowable analytical errors. Clin Chem1989;35:284-8.

Mahfoud SW. Crowding and preselection revisited. In Männer R, Manderick B (eds). Parallel problem solving from nature, 2. Amsterdam: Elsevier Science Publishers B.V., 1992:37-46.

Marse K, Roberts SD. Implementing a portable FORTRAN uniform (0, 1) generator. Simulation 1983;41:135-9.

Packard NH. A genetic learning algorithm for the analysis of complex data. Complex Systems 4;1990:573-86.

Parvin CA. Estimating the performance characteristics of quality-control procedures when error persists until detection. Clin Chem 1991;37:1720-4.

The International Federation of Clinical Chemistry Committee on Standards. Provisional recommendation on quality control in clinical chemistry. Clin Chim Acta 1975:63: F25-38.

Westgard JO, Barry PL. Cost-effective quality control: managing the quality and productivity of analytical processes. Washington, DC: AACC Press, 1986:33-91, 190-191, 195-217.

Westgard JO et al. A multi- rule Shewhart chart for quality control in clinical chemistry. Clin Chem 1981;27:493-501.

Westgard JO, Groth T. Design and evaluation of statistical control procedures: Applications of a computer "quality control simulator" program. Clin Chem 1981;27:1536-45.

Wolfram S. Cellular automata and complexity. Reading, Massachusetts: Addison-Wesley. 1994: 516-22.